%% file: main.tex
\documentclass[conference]{IEEEtran}
\IEEEoverridecommandlockouts
\usepackage{cite}
\usepackage{amsmath,amssymb,amsfonts}
\usepackage{algorithmic}
\usepackage{graphicx}
\usepackage{textcomp}
\usepackage{xcolor}
\def\BibTeX{{\rm B\kern-.05em{\sc i\kern-.025em b}\kern-.08em
    T\kern-.1667em\lower.7ex\hbox{E}\kern-.125emX}}

\usepackage{hyperref}
\usepackage[utf8]{inputenc}
\usepackage{amsmath} 
\usepackage{algorithm}
\usepackage{amsfonts}
\usepackage{algorithmic}
\usepackage{subfig}
\usepackage{xcolor}
\usepackage{colortbl}
\usepackage{float, enumitem}
\usepackage{pifont}
\usepackage{multirow}
\usepackage{diagbox}
\usepackage{microtype}
\usepackage{booktabs}

\usepackage{placeins}
\usepackage{array}
\usepackage{makecell}
\usepackage{textcomp}
\usepackage{tikz}

\newcommand\copyrighttext{
\small{
978-1-7281-8003-8/20/\$31.00
\textcopyright 2020 IEEE. Personal use of this material is permitted. Permission from IEEE must be obtained for all other uses, in any current or future media, including reprinting/republishing this material for advertising or promotional purposes, creating new collective works, for resale or redistribution to servers or lists, or reuse of any copyrighted component of this work in other works.}}
\newcommand\copyrightnotice{%
\begin{tikzpicture}[remember picture,overlay]
\node[anchor=south,yshift=10pt] at (current page.south) {\fbox{\parbox{\dimexpr\textwidth-\fboxsep-\fboxrule\relax}{\copyrighttext}}};
\end{tikzpicture}%
}

\title{Data Poisoning Attacks on Regression Learning and Corresponding Defenses}

\begin{document}

\author{\IEEEauthorblockN{Nicolas M\"uller}
\IEEEauthorblockA{\textit{Fraunhofer AISEC} \\
Garching near Munich, Germany \\
nicolas.mueller@aisec.fraunhofer.de}
\and
\IEEEauthorblockN{Daniel Kowatsch}
\IEEEauthorblockA{\textit{Fraunhofer AISEC} \\
Garching near Munich, Germany \\
daniel.kowatsch@aisec.fraunhofer.de}
\and
\IEEEauthorblockN{Konstantin B\"ottinger}
\IEEEauthorblockA{\textit{Fraunhofer AISEC} \\
Garching near Munich, Germany \\
konstantin.boettinger@aisec.fraunhofer.de}
}

\maketitle
\copyrightnotice

\begin{abstract}
Adversarial data poisoning is an effective attack against machine learning and threatens model integrity by introducing poisoned data into the training dataset. 
So far, it has been studied mostly for classification, even though regression learning is used in many mission critical systems (such as dosage of medication, control of cyber-physical systems and managing power supply). 
Therefore, in the present research, we aim to evaluate all aspects of data poisoning attacks on regression learning, exceeding previous work both in terms of breadth and depth. 
We present realistic scenarios in which data poisoning attacks threaten production systems and introduce a novel black-box attack, which is then applied to a real-word medical use-case.
As a result, we observe that the mean squared error (MSE) of the regressor increases to 150 percent due to inserting only two percent of poison samples. 
Finally, we present a new defense strategy against the novel and previous attacks and evaluate it thoroughly on 26 datasets.
As a result of the conducted experiments, we conclude that the proposed defence strategy effectively mitigates the considered attacks.
\end{abstract}

\section{Introduction}
Regression learning is increasingly used in mission-critical systems: In medicine for the development of pharmaceuticals \cite{ekins2019exploiting, Kuchler2019},
in the financial sector for predictive analysis, such as managing hedge funds \cite{Wigglesworth_2019,Porzecanski_2019} and cash forecasting \cite{Handelsblatt_2018},
as well as for predictive maintenance \cite{PdM} and quality control \cite{Juarez_2019}.

As we rely more and more on these systems, researchers find that they are vulnerable to malicious attacks.
Two strains of attacks can be distinguished.
Attacks at test time (evasion) and attacks at training time (poisoning) \cite{8406613}.
In this work, we focus on the latter. Poisoning attacks introduce a small fraction of 'poisoned samples' into the training process, which maximally 'confuses' the learner and causes either a denial of service (i.e., renders the model useless with respect to its original intent), or introduces a 'backdoor', which gives the attacker control over the model at test time.

These attacks (and corresponding defenses) have been studied by the scientific community in detail for classification learning \cite{Chen2018, Xiao2015, Biggio2012, Shafahi2018, Steinhardt}.
However, there is almost no research on adversarial poisoning attacks (and corresponding defenses) for regression learning. We find only a single contribution \cite{Jagielski2018} for regression learning, even though regression is used in many mission-critical systems as described above.

As an example, consider the following medical scenario.
The blood thinner Warfarin has a very small therapeutic window;
too high dosage leads to bleeding, while too low dosage leads to clotting  \cite{WikiWarfarin}.
Machine learning can help in estimating the correct dosage, and pharmaceutic companies provide appropriate datasets \cite{PharmGKB_2019} on which regression learning has been successfully applied \cite{Sharabiani_Bress_Douzali_Darabi_2015, Ma_Wang_Gao_Wang_Khalighi_2018}.

However, such an estimation is susceptible to data poisoning attacks:
A malicious entity may introduce a very small percentage of 'poisoned samples' into the dataset. Motives for such an attack can be manifold:
Personal motives (a malignant doctor, underpaid caregiver or simply a psychopath nurse \cite{NYT2019}), financial motives (one company damaging another company's reputation, or an individual betting on the crash of some company's stock value, similar to \cite{Rogers_2018}), or even political or terrorist motives.

Such poisoning attacks are not just theoretical threats:
Since even small data poisoning can have significant effect, such an attack is practicable even by an individual. 
Related work has shown that data poisoning is feasible and has already been observed in real-life scenarios\cite{Shafahi2018, Chen2018}.
To address this issue, we 
\begin{itemize}[topsep=0pt]
    \itemsep-0.2em
    \item show the harmfulness of data poisoning in regression by means of studying the problem of Warfarin dose prediction,
    \item present a new black-box attack which exceeds previous state-of-the-art, and for the first time evaluate poisoning attacks on nonlinear regression learners,
    \item present an improvement to previously suggested defenses which consistently outperforms the baseline,
    \item thoroughly evaluate our attack and defense on 26 datasets and state-of-the-art regression learners, i.e. Neural Networks, Kernel SVR, and Kernel Regression,
    % \item and open-source all code to support full reproducibility at \url{https://github.com/ANONYMIZED},
    \item and publish all source code and experiments to enable full reproducibility.
\end{itemize}

\section{Case Study: Warfarin Dose Estimation}
\label{s:case_study_warfarin}

In order to demonstrate the effect of even small fractions of poison samples, we examine the medical use case of Warfarin dose prediction in this section.
The International Warfarin Pharmacogenetics Consortium \cite{PharmGKB_2019}, a group of pharmocogenetic research centers, have created the IWCP dataset (\emph{Warfarin dataset}).
 It is the joint effort of 59 contributors, resulting in an average contribution of $1.7\%$ of the data per member.
 Based on this dataset, models have been developed which predict the therapeutic dose of Warfarin for a patient~\cite{Sharabiani_Bress_Douzali_Darabi_2015,Ma_Wang_Gao_Wang_Khalighi_2018}.

We use a new black-box poisoning attack(to be detailed in Section~\ref{ss:flip}) and add $2\%$ poison data to the Warfarin dataset, which is about as much as the average IWPC contributor did.
Table \ref{tab:warfarin_mae} shows the Mean Absolute Error (MAE) of different models after training on this dataset.
In the absence of poison samples, the MAE of models like Lasso, Elastic Net and Ridge is around $8.50$, which is comparable to state of the art \cite{Ma_Wang_Gao_Wang_Khalighi_2018}.
When adding just $2\%$ of data poisoning, the median error increases to 11.07 (a 29 percent increase). 
This has tangible effect on the patients:
The rate of acceptable doses\footnote{Following \cite{Ma_Wang_Gao_Wang_Khalighi_2018}, we measure the percentage of patients whose predicted dose of Warfarin is within $20\%$ of the true therapeutic dose. This is referred to as an \emph{acceptable dose}.} decreases by $21\%$.
% We will revisit this use case in more detail in Section~\ref{s:warfarin_revisited}, where we give more technical details and also show how to defend against such attacks.

\begin{table}[h]
    \centering
    \input{warfarin_results.tex}

    \caption{Mean absolute error (MAE) of different regression models when poisoning the \emph{Warfarin} dataset. The first column shows the MAE when no data poisoning is present.
    The second column shows the MAE when $2\%$ poison samples are introduced, with the relative change indicated by the third column.
    The poisoning strongly affects the amount of patients who receive an \emph{acceptable dosage} of Warfarin (fourth column).}
    \label{tab:warfarin_mae}
\end{table}

\section{Related Work}
In this section, we give a short overview on existing literature on data poisoning.
\subsection{Poisoning Attacks in Classification}
Early work on data poisoning attacks against classifiers include \cite{Biggio2012,Xiao2015}, which use the Karush-Kuhn-Tucker conditions to find optimal poisoning samples against linear models.
\cite{Biggio2012} first develops a poisoning attack against SVMs. \cite{Xiao2015} considers the security of feature selection against poisoning attacks and adapts the approach for LASSO, Ridge Regression and Elastic Net.
In both scenarios the attacker attempts to increase the test error and, thus, decrease the overall performance of the classifier.

\cite{Munoz-Gonzalez:2017:TPD:3128572.3140451} are the first to extent data poisoning to the multi-class scenario, which allows for targeted attacks. Instead of the Karush-Kuhn-Tucker conditions, they use back-gradient optimization to generate the first poison samples for neural networks in an end-to-end fashion without the requirement of a surrogate model. While these first results indicate a higher resilience of deep neural networks against availability attacks, \cite{Munoz-Gonzalez:2017:TPD:3128572.3140451} also shows the effectiveness of this approach against simpler models like MLPs with a single hidden layer.

\cite{Shafahi2018} build upon the work of \cite{Munoz-Gonzalez:2017:TPD:3128572.3140451} and demonstrate reliable clean-label attacks, in which the attacker can control the input data $\mathbf{X}$, but not the corresponding labels $\mathbf{y}$.
The attackers objective is to achieve misclassification of a certain instance as another class at test time.
For a transfer learning scenario, they show, that a single poison sample is capable of successfully poisoning a classifier.
For end-to-end learning settings they develop a watermarking approach to poisoning.

\subsection{Poisoning Attacks in Regression}
Data poisoning has so far been examined almost exclusively for classification learning.
For regression learning, there is work only by Jagielski et al. \cite{Jagielski2018}. They build upon work by Xiao et al. \cite{Xiao2015}, who introduce a gradient-based optimisation attack for linear classifiers such as Lasso, Ridge Regression and Elastic Net for feature selection.
Jagielski et al. \cite{Jagielski2018} use the same approach for the same models, but interpret the model's decision surface as a predictor for the continuous target variable, yielding a poisoning attack for linear regression.
Additionally, they introduce a non-gradient based attack, plus a defense called \emph{Trim} and evaluate it on three datasets.
Their approach in evaluating the defense is, however, not applicable in practice, since they use an oracle to determine the defense's hyper parameters. More specifically, they assume they know the fraction $\epsilon$ of poisoned samples in the dataset of size $n$, which is generally unknown.

Nonlinear regressors such as Kernel Ridge, Kernel SVM and Neural Networks have, to the best of our knowledge, not yet been examined in the context of adversarial poisoning.
This may be because the attack presented in \cite{Xiao2015} is not applicable to nonlinear learners.

\section{Poisoning Attacks in Regression}
In this section, we present our threat model, previously suggested attacks and our proposed and improved attack.
A thorough evaluation on 26 datasets is given in Section~\ref{s:eval}.

\subsection{Threat Model}
We consider a realistic attack scenario where the attacker has only limited capabilities, such as for example a malicious individual could have. 
Specifically, we consider black-box attacks where 
1) the attacker knows nothing about the model (not even what kind of regressor is used),
2) the attacker does not have access to the training dataset $(\mathbf{X}, \mathbf{y})$, but only to a smaller substitute dataset $(\mathbf{X}^{sub}, \mathbf{y}^{sub})$, and
3) where the attacker is capable of fully controlling the $\epsilon n$ data samples he contributes to the dataset. He is not able to manipulate the rest of the data.

As indicated in the introduction and Section \ref{s:case_study_warfarin}, the possibility of introducing small amounts of poison data into the dataset is highly realistic.
If the data are crawled and collected automatically, malicious instances just need to be placed where the crawler can find them \cite{Shafahi2018, Perdisci2006}.
If data are collected manually, the ability to poison a dataset is proportional to an individual's contribution to the dataset.
As detailed in Section \ref{s:case_study_warfarin}, the Warfarin dataset is collected by 59 individuals; thus, an average contribution constitutes about $2\%$ of the dataset.
We show that this amount of poisoning is sufficient to effectively poison the dataset (c.f. Section~\ref{s:case_study_warfarin}, ~\ref{ss:eval_flip}, and \ref{s:warfarin_revisited}).

\subsection{Related Poisoning Attacks in Regression}\label{ss:related_work_attacks}
\cite{Jagielski2018} present both a white box and a black-box attack on regression learning.
% However, both attacks cannot be applied in a black-box setting.
In this section, we present these attacks and and their limitations.

\subsubsection{Related White Box Attacks}
Deriving from \cite{Xiao2015}, a white-box attack on linear regressors is presented in \cite{Jagielski2018}.
The attacker's objective is formulated as a bilevel optimization problem as follows:
\begin{align}
    \arg \max_{\mathbf{D_p}} & \hspace{10pt} \mathcal{W}(\mathbf{D'}, \theta_p) \label{e:1} \\
    s.t. & \hspace{10pt}  \theta_p \in \arg \min_\theta \mathcal{L}(\mathbf{D_{tr}} \cup \mathbf{D_p}, \theta) \label{e:2}
\end{align}
Equation~\ref{e:2} is the usual minimization of the model loss $\mathcal{L}$ during the fitting of a model on both the clean training dataset $\mathbf{D_{tr}}$ and the poisoned dataset $\mathbf{D_p}$.
This yields an optimal set of weights $\theta_p$.
This is called the 'inner optimisation'.
 
Equation \ref{e:1} refers to maximising the attacker's objective $\mathcal{W}$ with respect to some test set $\mathbf{D'}$, using the model's weights as determined by Equation \ref{e:2}.
Minimizing Equation \ref{e:1} depends on the solution of Equation \ref{e:2}, which is why it is considered a bilevel optimization problem.
% The difficulty of this problem lies in the fact that the attacker has to determine how introduced data will affect model weights during training.
This is a hard problem: The attacker has to determine how the points they introduce in the dataset will change the model weights during training.
\cite{Xiao2015, Jagielski2018} solve this using the Karush-Kuhn-Tucker (KKT) conditions as a set of conditions which they assume remain satisfied when a given poison sample $\mathbf{x_c}$ is introduced.
They then proceed in  solving a linear system, and, thus, derive the gradients.
 
This approach is not feasible in deep neural networks \cite{Munoz-Gonzalez:2017:TPD:3128572.3140451}, since the time required for solving the linear system is in $O(p^3)$, where $p$ is the number of parameters in the model.
Since even small, commonly used pretrained models have a few million parameters \cite{BibEntry2019Sep}, the computation is not feasible.
Even with simplifying assumptions or a sufficiently small number of parameters, this approach still requires an exact solution to the optimization problem, which, in general, can not be obtained.
For a more detailed analysis we refer to \cite{Munoz-Gonzalez:2017:TPD:3128572.3140451}.
% Additionally, this white box approach requires access to the gradients of the model.
% In our scenario we assume a black-box scenario where the attacker has no knowledge of the model, 
% thus such a gradient-based attack is not practicable.

\subsubsection{Related Black-Box Attacks}
\cite{Jagielski2018} also present a black-box attack called \emph{StatP}.
This attack samples $\epsilon n$ points from a multivariate Gaussian distribution, where the corresponding mean $\mathbf{\mu}$ and co-variance matrix $\mathbf{\Sigma}$ are estimated as the mean and co-variance of the true dataset $\mathbf{D_{tr}}$.
Then, \emph{StatP} rounds the feature variables to the corners, queries the model and rounds the target variable to the opposite corner.
%and target values to the corners, where the corners are defined by the minimum and maximum values still in the feasibility domain $\gamma$.
The corners are defined as the minimum and maximum of the feasibility domain $\gamma$ of each variable.
Both features and target are scaled to $[0, 1]$, thus the feasibility domain is a hypercube $[0, 1]^{d+1}$ where $d$ is the number of features.
In summary, this attack creates a few isolated clusters of adversarial data, where both features and target take only extreme values of either $\gamma_{min}=0$ or $\gamma_{max}=1$.

This attack, however, still requires access to the trained black-box model, which may be unrealistic in a real-world scenario. Additionally, we find that while this attack is successful on linear models, it is unsuccessful when applied to non-linear models.
We show this empirically in Section~\ref{s:eval}, but give a brief explanation here:
Nonlinear learners (such as Neural Networks, Kernel SVR, and Kernel Regression) are able to accommodate both the poison points and the true data simultaneously.
This is because the poison data created by \emph{StatP} does not contradict the true data points, since true data points rarely have features in the corners of the feasibility domain.
This insight will motivate our proposed \emph{Flip} attack on nonlinear learners, which we present in the next section.

\subsection{Flip: A Black-Box Attack on Nonlinear Regressors}
\label{ss:flip}

\begin{algorithm}[t]
\caption{Flip attack}\label{alg:1}
\begin{algorithmic}[1]
\REQUIRE
    \STATE Substitute data $\mathbf{X}^{sub}, \mathbf{y}^{sub}$ of size $m$
    \STATE Number of poison points $\lceil \epsilon n \rceil$ to compute
    \STATE Feasibility domain $[\gamma_{min}, \gamma_{max}]$ of the target values
% \Statex
\FUNCTION{Flip}{}
\STATE $\mathbf{\Delta} \xleftarrow{} \emptyset$
\FOR{$i \in [1, ..., m]$}
    \STATE $\Delta_i \xleftarrow{}  \max( y^{sub}_i - \gamma_{min} , \gamma_{max} - y^{sub}_i )$
    \STATE $\mathbf{\Delta} \xleftarrow{} \mathbf{\Delta} \cup \Delta_i$
\ENDFOR
\STATE $T_\epsilon \xleftarrow{} t \in \mathbb{R}$ 
s.t. $t$ is the $\lceil \epsilon n \rceil$-th highest value of $\mathbf{\Delta}$
\STATE $I_\epsilon \xleftarrow{} \{ i \in [1, ..., m] \text{ s.t. } {d}_i >= T_\epsilon \text{ where } d_i \in \Delta\}$
\STATE $\mathbf{X_p} \xleftarrow{} \emptyset$, \hspace{5pt} $\mathbf{y_p} \xleftarrow{} \emptyset$
\FOR{$i \in I_\epsilon$}
    \IF{$y_i > \frac{1}{2}(\gamma_{max} - \gamma_{min})$}
        \STATE $ y_{p, i} \xleftarrow{} \gamma_{min}$
        \STATE $\mathbf{y_p} \xleftarrow{} \mathbf{y_p} \cup {y_{p, i}}$
    \ELSE
        \STATE $ y_{p, i} \xleftarrow{} \gamma_{max}$
        \STATE $\mathbf{y_p} \xleftarrow{} \mathbf{y_p} \cup {y_{p, i}}$
    \ENDIF
    \STATE $\mathbf{X_p} \xleftarrow{} \mathbf{X_p} \cup {X}^{sub}_i$
\ENDFOR
\STATE \textbf{return} $\mathbf{X_p}, \mathbf{y_p}$
\ENDFUNCTION
\end{algorithmic}
\end{algorithm}

Algorithm~\ref{alg:1} presents our proposed black-box attack called  \emph{Flip}.
This algorithm computes a set of adversarial poisoning points for any degree of poisoning $0 < \epsilon < 1$.
The attack is completely independent of the regressor model and only requires a substitute dataset $(\mathbf{X}^{sub}, \mathbf{y}^{sub})$ from the same domain as the training dataset $\mathbf{D_{tr}}$ and a feasibility domain of the target variables $[\gamma_{min}, \gamma_{max}]$.
The feasibility domain is necessary because we usually assume that only certain target variables are valid.
Other values are bound to raise suspicion, such as for example a room temperature of $-400$ degrees Celsius, or medical doses that are extremely high or low.

We now describe our attack.
After having initialised an empty set $\mathbf{\Delta}$ in line 2, we populate it in the following \emph{for loop} (line 3-6).
For each instance in the substitute dataset, we find the maximum of the distance to the lower or upper end of the feasibility domain, and save the results to $\mathbf{\Delta}$.
Then, in line 7 we find the $\lceil \epsilon n \rceil$-highest value $t \in \mathbb{R}$ in $\mathbf{\Delta}$.
This is used in line 8 to compute the indices of those points for which there is most potential to disturb.
Thus, the rational of line 7-8 is to find those points for which the target value is closest to either $\gamma_{min}$ or $\gamma_{max}$.
These values are the ones which can be maximally disturbed by shifting the target variable to the other side of the feasibility domain.
This is implemented in line 10-19, where we compute the poison set by retaining the feature values and 'flipping' the target value to the other side of the feasibility domain for the appropriate candidates as specified by $I_\epsilon$.
Finally, in line 20, we return the found poison data. 
% TODO Add actual link
% We publish an implementation at \url{https://github.com/ANONYMIZED}.
% TODO not sure if we want to do this

%\todo{Worst case complexity is in $O(mlog(m))$ because of finding $T_\epsilon$}

\section{Data Poisoning Defenses}\label{s:defenses}

In this section, we present defenses for adversarial data poisoning in regression.
First, we propose a set of requirements to make a defense applicable in practice.
Second, we evaluate existing defenses with respect to these requirements.
Finally, we present our improvement over the baseline.
A quantitative evaluation is given in Section~\ref{s:eval}, while a qualitative evaluation is presented in Section~\ref{s:warfarin_revisited}.
%Section~\ref{s:warfarin_revisited}.

\subsection{Requirements for Data Poisoning Defenses}
When creating a dataset such as \cite{PharmGKB_2019}, the defender does not know the degree of data  poisoning, if any. 
Put differently, while $\epsilon$ might be known to the attacker, it is unknown to the defender.
It is also entirely possible that no poisoning has happened, i.e. that $\epsilon = 0$.
Thus, the quality of a defense should not depend too much on a correct guess of $\epsilon$, and should not deteriorate the quality of an unpoisoned dataset.
With this requirement in mind, we proceed to present existing defenses and evaluate them against it.

\subsection{Related Defenses}\label{ss:related_defs}
Since there exists so little work on poisoning in regression, existing defenses are also few.
Two defenses from the domain of classification learning are presented in \cite{Steinhardt}. The \emph{Sphere} defense first computes centroids in the poisoned data, and then removes points outside a spherical radius around the centroids.
The \emph{Slab} defense 'projects points onto the line between the centroids and then discards points that are too far away' \cite{Steinhardt}.
However, \cite{Steinhardt} themselves note that these defenses may leave datasets vulnerable, and present an example based on the IMDB classification dataset where both \emph{Sphere} and \emph{Slab} fail.
For this reason, we do not consider them to be viable.
%, which is why we do not consider them to be viable.

\begin{algorithm}
\caption{Trim defense, as proposed in \cite{Jagielski2018}}\label{alg:2}
\begin{algorithmic}[1]
\REQUIRE
    \STATE Poisoned dataset $\mathbf{D_{tr}} \cup \mathbf{D_p}$ of combined size $n$
    %\Statex Feasibility domain $[\gamma_{min}, \gamma_{max}]$ of the target values
    \STATE Some model loss $\mathcal{L}$
    \STATE Estimated fraction of poison points $ \hat{\epsilon}$
\FUNCTION{Trim}{}
\STATE $\mathcal{I}^{(0)} \leftarrow $ a random subset of indices with size $n\frac{1}{1+\hat{\epsilon}}$
\STATE $\theta^{(0)} \leftarrow \arg \min_\theta \mathcal{L}(D^{\mathcal{I}^{(0)}}, \theta)$ 
\STATE $i \leftarrow 0$
\WHILE{True}
    \STATE $i \leftarrow i + 1$
    \STATE $\mathcal{I}^{(i)} \leftarrow $ subset, size $n\frac{1}{1+\hat{\epsilon}}$, min. $\mathcal{L}(D^{\mathcal{I}^{(i)}}, \theta^{i-1})$
    \STATE $\theta^{(i)} \leftarrow \arg \min_\theta \mathcal{L}(D^{\mathcal{I}^{(i)}}, \theta)$ 
    \STATE break if some convergence condition is met
\ENDWHILE
\STATE \textbf{return} $D^{\mathcal{I}^{(i)}}$
\ENDFUNCTION
\end{algorithmic}
\end{algorithm}

The state-of-the-art defense is \emph{Trim} \cite{Jagielski2018}, for which we give pseudo-code in Algorithm~\ref{alg:2}.
It is an iterative algorithm, which first fits a regressor to a subset of the poisoned data, and then iteratively calculates the error between the regressor's prediction on the \emph{train} set and the \emph{train} targets.
It refits the regressor on those points with the smallest error, and repeats until a convergence criteria is met.
Finally, it returns the points with the smallest error as a 'cleaned' dataset.
The number of points to fit on, and conversely, the number of points to discard, is determined by a supplied parameter $\hat{\epsilon}$, the assumed degree of poisoning.
If $\hat{\epsilon} = \epsilon$, the defense has been shown to work very well \cite{Jagielski2018}. However, this is not a realistic scenario, since the defender does not know $\epsilon$.

\begin{figure}[!ht]
    \centering
    \includegraphics[width=0.90\linewidth]{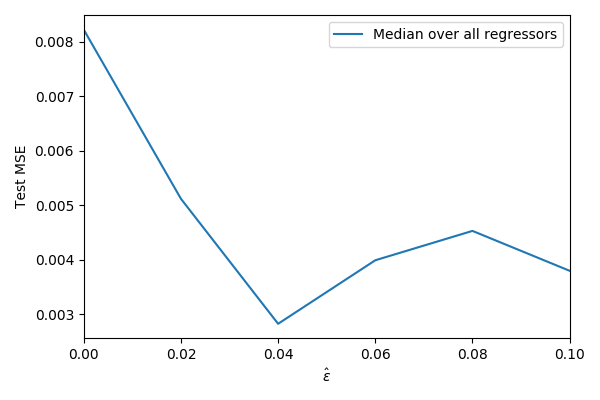}
    \caption{Three real-world datasets (\emph{Warfarin}, \emph{Loan} and \emph{Housing} \cite{Jagielski2018}) are poisoned with the \emph{Flip} attack, where $\epsilon = 0.04$. We then apply the \emph{Trim} defense with different values of $\hat{\epsilon}$, and train a regressor on the resulting dataset. Finally, we plot the test MSE against $\hat{\epsilon}$, averaged over all datasets and regressors. Observe that the \emph{Trim} defense is highly dependent on the correct choice of $\hat{\epsilon}$ (in this case $0.04$): If $\hat{\epsilon}$ is estimated too low, poison points remain in the training data, skewing the regressor and causing high test loss. If $\hat{\epsilon}$ is estimated too high, legitimate points are removed with the poisoned points. This loss in training data causes the regressor to learn a distribution different from the test distribution, which also incurs higher test loss. Thus, it is important to accurately estimate the degree of poisoning $\hat{\epsilon}$.}
    \label{fig:defense_and_eps}
\end{figure}

Consider Figure~\ref{fig:defense_and_eps}, where we poison three real-world datasets from \cite{Jagielski2018}, including the \emph{Warfarin} dataset, with a poison data fraction of $\epsilon = 0.04$.
Then, for each dataset, we clean it using the \emph{Trim} defense, where we supply $\hat{\epsilon} \in [0.00, 0.02, 0.04, 0.06, 0.08, 0.10]$ (i.e. we clean the poisoned dataset with different estimates $\hat{\epsilon}$ to quantify the effect of $\hat{\epsilon}$ on \emph{Trim}).
On the resulting data (which is partially or fully free from poison samples, depending on $\hat{\epsilon}$), we train a regressor and calculate the MSE on a separate test set.
Then, we average the MSE over all datasets and plot the median of the regressors against $\hat{\epsilon}$.
%The MSE is then averaged over all datasets and regressors and plotted against $\hat{\epsilon}$.
% REMOVED BECAUSE OF SPACE
%For a plot showing the effects on each regressor individually, we refer to Figure~\ref{fig:defense_and_eps_all} in the Supplementary Material.
%This MSE is then plotted against $\hat{\epsilon}$.

We make one key observation: The effectiveness of \emph{Trim} highly depends on the correct choice of $\hat{\epsilon}$.
Selecting $\hat{\epsilon}$ below the actual degree of poisoning results in not all poison samples being removed and, thus, in an increase of the test MSE of an regressor.
Selecting $\hat{\epsilon}$ above the actual degree of poisoning results in pristine data being removed, which might also remove relevant structure/information contained in the dataset and, as a result, also increase test MSE.
Therefore, a better selection strategy than blind overestimation of $\hat{\epsilon}$ is required.

\subsection{The Iterative Trim Defense}\label{ss:itrim}
\begin{algorithm}
\caption{iTrim defense}\label{alg:3}
\begin{algorithmic}[1]
\REQUIRE
    \STATE Poisoned dataset $\mathbf{D_{tr}} \cup \mathbf{D_p}$ 
    %\Statex Feasibility domain $[\gamma_{min}, \gamma_{max}]$ of the target values
    \STATE Some model loss $\mathcal{L}$
    \STATE Maximum estimated poisoning rate $\epsilon_{max}$
    \STATE Number of runs $r$
    \STATE Threshold $t$
\FUNCTION{iTrim}{}
\STATE $ I \leftarrow \Big \{  \epsilon_{max} \frac{j}{r-1} $ s.t. $j \in \{0, ..., r-1\} \Big \} $
\FOR{$i \in I $}
    \STATE $\mathbf{D}^{(i)} \leftarrow trim(\mathbf{D_{tr}} \cup \mathbf{D_p},  \mathcal{L}, \hat{\epsilon} = i) $
    \STATE $L^{(i)} \leftarrow \min_\theta \mathcal{L}(\mathbf{D}^{(i)}, \theta)$
\ENDFOR
\STATE $\epsilon_{opt} \leftarrow \min \{ i \in I $ s.t. $ | L^{(i)} - L^{(i-1)} | < t \} $ 
\STATE \textbf{return} $trim(\mathbf{D_{tr}} \cup \mathbf{D_p},  \mathcal{L}, \hat{\epsilon} = \epsilon_{opt})$
\ENDFUNCTION
\end{algorithmic}
\end{algorithm}

As shown in the last subsection, the \emph{Trim} defense has the potential to accurately remove poison samples from a given dataset, provided that $\hat{\epsilon}$ is chosen correctly, but over- or underestimating $\hat{\epsilon}$ significantly decreases test performance.
From this result stems the motivation for our proposed \emph{Iterative Trim} defense (\emph{iTrim}).
This defense enhances \emph{Trim} by an iterative search for the best $\hat{\epsilon}$.
In this section, we present this algorithm and our proposal for selecting the ideal value for $\hat{\epsilon}$.
In Section~\ref{s:eval}, we show empirically on 26 datasets that \emph{iTrim} can be applied under realistic conditions to poisoned data, and reliably identifies and removes the poisoned data.

\subsubsection{Algorithm Description}
Algorithm~\ref{alg:3} details the \emph{iTrim} defense.
It takes as arguments the poisoned dataset $\mathbf{D_{tr}} \cup \mathbf{D_p}$, a loss $\mathcal{L}$, and three scalar hyper parameters.
The first, $\epsilon_{max}$, is an estimate of the maximum possible poisoning rate.
This hyper parameter can be chosen arbitrarily large without impacting the defense's result, but if chosen correctly will improve run time.
The second hyper parameter specifies the number of runs $r$.
This hyper parameter does not have too much influence on the algorithm's performance; it influences together with $\epsilon_{max}$ which values of $\hat{\epsilon}$ will be tried.
The final hyper parameter, the threshold $t$, does have impact on the algorithm's performance, and we will discuss how to chose it later on.

\emph{iTrim} starts by calculating a set $I$ of possible candidates $\hat{\epsilon}$ (line 2). The hyper parameters $\epsilon_{max}$ and $r$ define the right bound and the number of points, respectively.
Then, for each candidate, calculate the cleaned dataset $\mathbf{D}^{(i)}$ using Trim, train the regressor and obtain the corresponding train loss $L^{(i)}$ (lines 3 - 4).
Finally, the optimal value for $\hat{\epsilon}$ is found when the error in train loss between two consecutive losses $ L^{(i)} - L^{(i-1)} $ first undercuts some threshold $t$ (line 7).
The dataset is cleaned using \emph{Trim} with this estimate, and the result is returned (line 8).

\subsubsection{Poison Rate Selection}
Before we give an intuition for our algorithm and show the reasoning for our selection criterion, we shortly address validation approaches to finding $\hat{\epsilon}$.
As already mentioned, $\hat{\epsilon}$ is a hyper parameter of \emph{Trim}. In machine learning, a common approach to finding hyper parameters are validation schemes, e.g. cross validation.
But for this approach to work, we require a clean validation dataset.
Since we only have a single dataset, we have to assume that any validation split will contain poisoned instances, rendering conventional validation approaches unsuited for finding hyper parameters in this setting.

\begin{figure}
    \centering
    \includegraphics[width=0.90\linewidth]{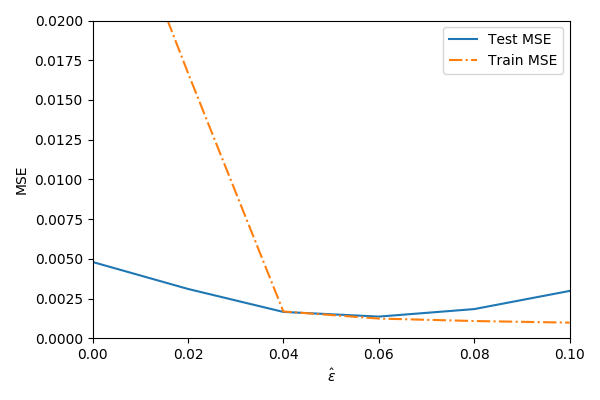}
    \caption{Applying \emph{Trim} to the \emph{Warfarin} dataset, poisoned  with $\epsilon = 0.04$. 
    The orange dashed line shows the averaged train loss for different estimations of poisoning $\hat{\epsilon}$, using \emph{KernelRidge} regression. There is high train loss for $\hat{\epsilon} = 0$ and $0.02$, because in these cases, not all poison points in the dataset can be removed. 
    Once all adversarial poison data is removed ($\hat{\epsilon} = 0.04$, the delta in train loss decreases by several orders of magnitude, approaching zero. The line in blue shows the test error for the same setup. Note that the best $\hat{\epsilon}$ is indeed characterized by sudden change of train loss, as indicated in the left figure.
    Also note that for $\hat{\epsilon} > 0.04$, \emph{Trim} starts to remove legitimate data, which deteriorates test performance.}
    \label{fig:iTrim_warfarin}
\end{figure}

Thus, we now proceed to explain our iterative approach to finding $\hat{\epsilon}$:
Consider Figure~\ref{fig:iTrim_warfarin}, where we apply \emph{Trim} to the \emph{Warfarin} dataset poisoned with $\epsilon = 0.04$.
The orange dashed line shows the train loss for different candidate values $\hat{\epsilon} \in [0.00, 0.02, 0.04, 0.06, 0.08, 0.10]$.
Note that for the correct estimation of poisoning degree $\hat{\epsilon} = 0.04$, the train loss becomes almost zero, decreasing several orders of magnitude compared to $\hat{\epsilon} = 0.00$.
Further increasing $\hat{\epsilon}$ still decreases the train MSE, but only insignificantly. % compared to $\hat{\epsilon}<0.04$.
Thus, the train loss can be approximated by two straight lines, joined at a distinctive kink where $\hat{\epsilon} = \epsilon$.
Figure~\ref{fig:iTrim_housing_loan} in the Supplementary Material shows this for other real-world datasets.
%, but the gradient below $\epsilon$ is noticeably steeper than above $\epsilon$.
We can understand this kink as the point where the dataset ceases to contain data which incurs extremely high train loss - in other words, where all adversarial poison data have been removed.
This assumption is supported by the blue line in Figure~\ref{fig:iTrim_warfarin}, which shows the test MSE for the same \emph{Kernel Ridge} regressor trained on the thusly cleaned datasets.
For $\hat{\epsilon} = 0.04$, the test loss is minimal.
For $\hat{\epsilon} > 0.04$, \emph{Trim} starts to remove legitimate data (since all poison data have been removed), which is why test performance deteriorates.
Section~\ref{s:eval} will verify this empirically.

Based on the insight that the train loss can be approximated by two straight lines which intersect at $\epsilon$, we develop our selection criterion for $\hat{\epsilon}$.
We define $t$ as the maximum absolute gradient of the straight line where $\hat{\epsilon} > \epsilon$ (i.e. the slope of the orange dashed line on the 'right' side of the graph, where all poison data have been removed).
We will refer to this straight as the \emph{normal straight}.
Then $| L^{(i)} - L^{(i-1)} |$ is used to approximate the gradient of the straight for each subinterval.
The division by the length of the interval is omitted since all intervals are equidistant.
We choose $\hat{\epsilon}$ as the first candidate so the estimated straight is normal (i.e. $| L^{(i)} - L^{(i-1)} | < t$).

\subsubsection{Threshold Selection}
\emph{iTrim} is dependent on an appropriate choice of the threshold $t$.
If $t$ is vastly too large, poison points are left in the dataset.
If $t$ is too small, \emph{iTrim} deteriorates to \emph{Trim}, and starts removing non-poison points.
However, we find that there is rather a large window of appropriate values of $t$.
This is because 1) we apply feature/target scaling to $[0, 1]$, and 2) the difference in train loss we observe once all poisoned points are removed is dramatic (c.f. Figure~\ref{fig:iTrim_warfarin}).
Based on our evaluation on 26 datasets (c.f. Section~\ref{ss:data_sets}), we find empirically that choosing values between $0.05$ and $0.0001$ performs comparably, and thus decide on a threshold $t = 0.001$.
In summery, we find that:
\begin{itemize}
    \itemsep0em
    \item \emph{Trim} lacks a mechanism to find a good estimate for the percentage $\hat{\epsilon}$ of poisoned points in the dataset.
    \item Over- and under estimation of $\hat{\epsilon}$ deteriorate the dataset to be cleaned.
    \item Appropriate values for $\hat{\epsilon}$ can be found via \emph{iTrim}.
\end{itemize}

\section{Empirical Evaluation}\label{s:eval}
In this section, we evaluate our attack and defense algorithms against 26 datasets.
We show that we 1) can reliably poison nonlinear and linear models while assuming a realistic black-box threat model, and 2) defend against this attack better than previously suggested defenses.

\subsection{Experimental Setup}
We try to make our experiment as general and realistic as possible. First, we split each of the 26 datasets into a randomly drawn \emph{substitute set} of size $0.25$, a \emph{train set} of size $0.75 * 0.8$ and a \emph{test set} of size $0.75 * 0.2$.
For each combination of the 26 \emph{substitute} datasets and $\epsilon \in [0.00, 0.02, 0.04, 0.06, 0.08, 0.10]$, we create a poisoned dataset using the respective attack, which we append to the corresponding \emph{train set} and shuffle.
This results in $6*26 = 156$ combinations of \emph{train} dataset and poisoning rate.
This step does not depend on the regressors.
Then, for each regressor and each of the 156 \emph{poisoned train datasets}, we perform Cross-Validated Grid Search to find suitable hyper parameters. Finally, for all 156 \emph{poisoned train  datasets} and both defenses (\emph{Trim} and \emph{iTrim}, we clean each of the 156 \emph{poisoned train datasets}.
We then train a regressor and measure test error on the \emph{test data} sets and report below.
Thus, in total we run $156 * 7 * 2 = 2184$ experiments ($7$ being the number of different regressors evaluated).
The experiments and source code are published to enable reproducibility\footnote{See \url{https://github.com/Fraunhofer-AISEC/regression_data_poisoning}}.

\subsection{Datasets and Regressors}\label{ss:data_sets}
For our experiments, we use 26 datasets:
Three datasets introduced in~\cite{Jagielski2018},
eight datasets from the GitHub repository \emph{imbalanced dataset} \cite{branco2019imbalanced},
and 15 datasets from the \emph{KEEL} regression repository~\cite{Alcala-Fdez2011}. 
Each dataset contains at least $1000$ data points.
For datasets where $n > 10000$, we randomly sample a subset $n = 10000$.
In keeping with~\cite{Jagielski2018}, we scale features and targets to $[0, 1]$.
See Table~\ref{tab:datasets} in the Supplementary Material for a detailed summary.
%Removed bc of space
%Refer to Table~\ref{tab:datasets} in the Supplementary Material for a detailed list.

We evaluate four linear models (HuberRegressor, Lasso, Ridge, Elastic Net) and three non-linear models (Neural Networks, Kernel Ridge with RBF kernel, and Support Vector Regressor with RBF kernel).
To the best of our knowledge, we are the first to evaluate poisoning attacks against non-linear regressors.

\subsection{Evaluation of \emph{StatP}}
In this section, we very briefly report the effectiveness of \emph{StatP} on non-linear regressors.
As detailed in \cite{Jagielski2018}, the attack is effective for linear regressors.
We find, however, that it is not effective when applied to non-linear learners.
For example, a Neural Network's MSE remains nearly unchanged (from $0.051$ to $0.055$) when poisoned with ten percent of poison samples created by \emph{StatP}. 
In the Supplementary Material, we elaborate this in more detail, and evaluate additional non-linear learners such as Kernel SVM and Kernel Ridge, which we find to behave similarly.

\subsection{Evaluation of \emph{Flip}}\label{ss:eval_flip}

\begin{figure}
    \centering
    \includegraphics[width=0.90\linewidth]{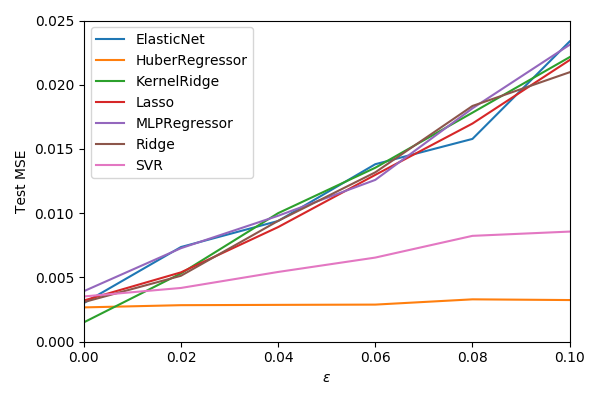}
    \caption{Evaluation of our proposed \emph{Flip} attack.  
    This plot shows the MSE for different poison rates per regressor, averaged over all 26 datasets. Most regressors obtain a MSE of around $0.003$ when $\epsilon = 0$, and all but two deteriorate linearly as $\epsilon$ increases.
    Figure~\ref{fig:attack_results_non_averaged} in the Supplementary Material shows the same results, but for each dataset individually.
    }
    \label{fig:flip_attack_all}
\end{figure}

In this section, we present the results when evaluating our proposed \emph{Flip} attack against 26 datasets and seven regressors.
Figure~\ref{fig:flip_attack_all} shows the performance of the \emph{Flip} attack, averaged over all datasets.
Figure~\ref{fig:attack_results_non_averaged} in the Supplementary Material shows results per dataset.

The attack is highly effective:
When adding only $4\%$ of poison data, the MSE of most regressors doubles compared to the non-poisoned case.
We observe that all models seem equally susceptible to our attack, with the exception of the Huber Regressor and Support Vector Regressor, which are designed to be outlier-resistant.

\subsection{Evaluation of \emph{Trim} and \emph{iTrim}}\label{ss:eval_trim}

In this section, we report the results when defending against the \emph{Flip} attack.
We report both the performance of the \emph{Trim} defense and our proposed \emph{iTrim} defense, and compare efficiency. We set both $\hat{\epsilon} = 0.14$ (for \emph{Trim}) and ${\epsilon}_{max} = 0.14$ (for \emph{iTrim}) to mimic the behaviour of a defender in a realistic scenario.
A defender would have to guess the percentage of poisoned data $\epsilon$, with a preference for overestimation rather than underestimation (as explained in Section~\ref{ss:itrim}).

\begin{figure}[]
    \centering
    \includegraphics[width=0.90\linewidth]{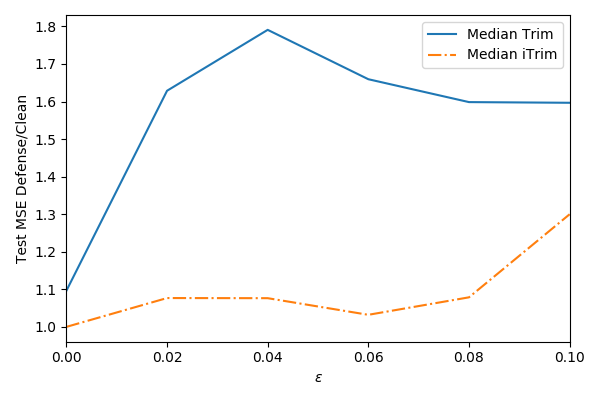}
    \caption{Evaluation of the \emph{Trim} and \emph{iTrim} defense. First, we poison the 26 datasets using \emph{Flip}.
    Then we apply the \emph{Trim} or \emph{iTrim} defense, and calculate the test MSE.
    Finally, we normalize by the \emph{baseline} MSE - the error obtained by a model trained on unpoisoned data.
    The resulting quotient represents the degree to which the defenses can negate the \emph{Flip} attack.
    We observe that \emph{iTrim} consistently outperforms \emph{Trim}.
    More detailed, non-averaged results are presented in Figure~\ref{fig:trim_itrim_def_all} and Figure~\ref{fig:defenses_results_non_averaged} in the Supplementary Material.
    }
    \label{fig:trim_itrim_def}
\end{figure}

\begin{figure*}
    \centering
    \includegraphics[width=0.40\linewidth]{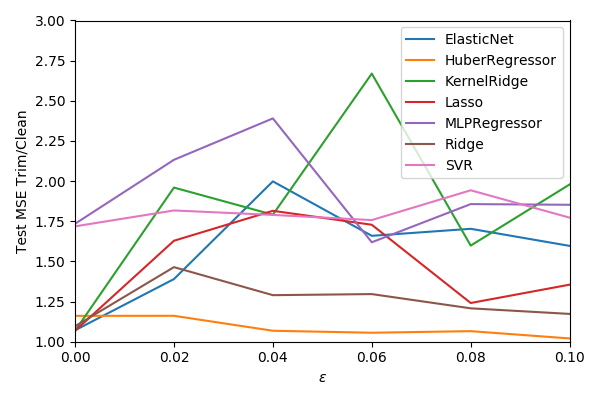}
    \includegraphics[width=0.40\linewidth]{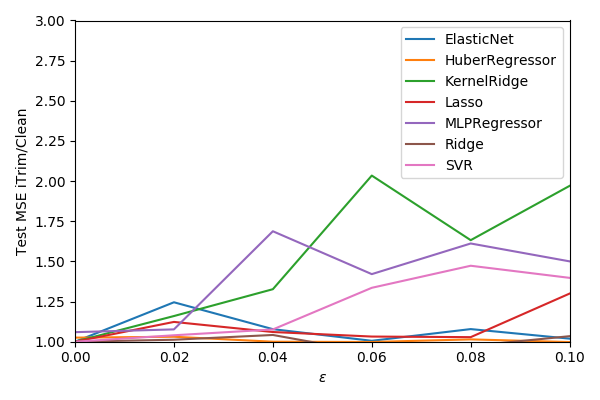}
    \caption{
    \emph{Trim} (left) vs. \emph{iTrim} (right), averaged by dataset.
    %Evaluation of the \emph{Trim} and \emph{iTrim} defense. 
    \emph{Left:} This plot shows the effectiveness of \emph{Trim} when defending against the \emph{Flip} attack, averaged over all 26 datasets. datasets are poisoned as indicated on the $x$ axis. Then, the \emph{Trim} defense is applied, the regressor is fitted to the dataset, and the resulting test MSE is compared against the test MSE on a clean, unpoisoned dataset. 
    We see that, depending on the regressor, datasets cleaned with \emph{Trim} still incur significant decrease in test performance. \emph{Right:} The same process is applied to \emph{iTrim}. We observe that test MSE is considerably decreased, especially for $\epsilon < 0.08$. The larger $\epsilon$, the more similar the two defences become. Additional information is provided by Figure~\ref{fig:defenses_results_non_averaged}.}
    \label{fig:trim_itrim_def_all}
\end{figure*}

\begin{figure*}
    \centering
    \includegraphics[width=0.40\linewidth]{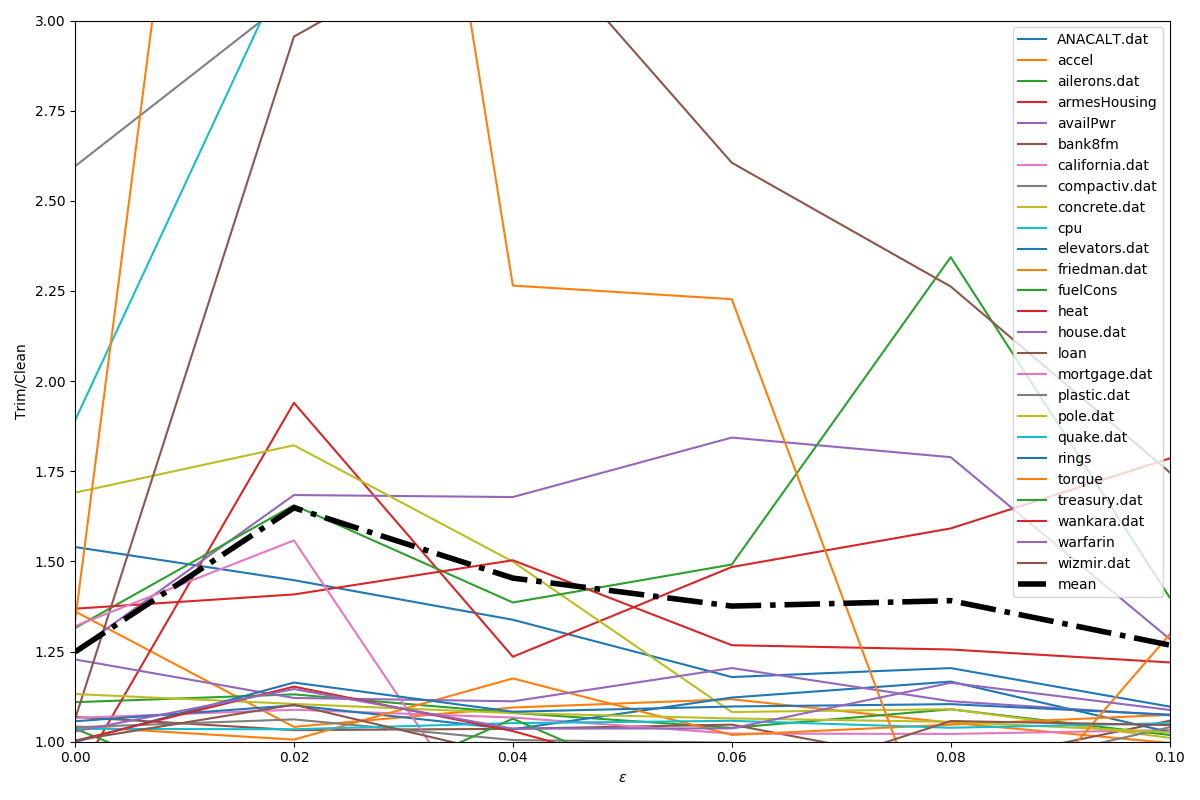}
    \includegraphics[width=0.40\linewidth]{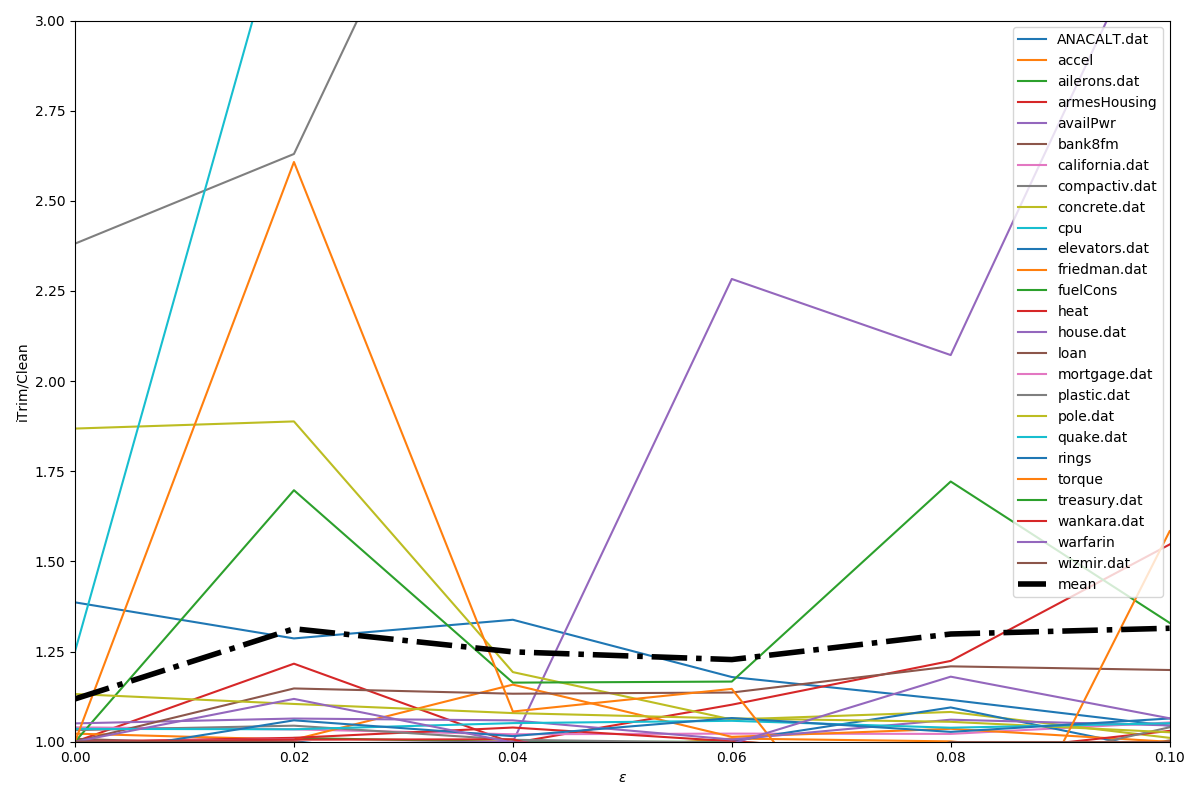}
    \caption{
    \emph{Trim} (left) vs. \emph{iTrim} (right), averaged by regressor.
    Similar to Figure~\ref{fig:trim_itrim_def_all}, this plot shows the increase in MAE when using a poisoned and subsequently defended dataset instead of a clean dataset during model training.
    The results are averaged over all seven regressors, but displayed individually per dataset.
    \emph{Left Image:} Increase in MAE when defending against a \emph{Flip} attack using \emph{Trim}.
    \emph{Right Image:} Increase in MAE when defending against a \emph{Flip} attack using \emph{iTrim}. Notice that \emph{iTrim} outperforms \emph{Trim} almost consistently (as seen by the overall lower MAE, shown in bold black)}
    \label{fig:defenses_results_non_averaged}
\end{figure*}

We proceed as follows:
With varying degrees of poisoning, we poison all 26 datasets using the \emph{Flip} attack.
Then, for each regressor, we clean (i.e. 'defend') the datasets using the \emph{Trim} as well as the \emph{iTrim} defense (separately).
We fit the regressor on the thusly obtained dataset, and compare the test error against a regressor trained on the 'clean' data.
Figure~\ref{fig:trim_itrim_def} shows the median of all regressors for \emph{Trim} (blue line) and \emph{iTrim} (orange dashed line)\footnote{
The Supplementary Material provides more details: Figure~\ref{fig:trim_itrim_def_all} presents the results for each regressor individually, while Figure~\ref{fig:defenses_results_non_averaged} depicts the results for all 26 datasets.
}.

We observe that both defenses are effective.
However, \emph{iTrim} achieves higher performance than \emph{Trim}, especially when there is a large discrepancy between $\epsilon$ and $\hat{\epsilon}$.
This is due to \emph{iTrim}'s capability of more accurately estimating the degree of poisoning.
% Two observations are noteworthy: First, \emph{iTrim} converges to \emph{Trim} when $\hat{\epsilon} \to \epsilon$. Second, 
Especially for $\hat{\epsilon} > \epsilon$, we see considerable improvement due to \emph{iTrims} more advanced estimate of $\epsilon$. 
Trim is also computationally feasible, despite its iterative approach: The average runtime for defending a given dataset was 1.6 minutes. A more detailed discussion can be found in the Supplementary Material.

\subsection{Runtime}
In this section, we detail the runtime of the \emph{iTrim} defense algorithm.
\emph{iTrim} calls the \emph{Trim} defense $r$ times, which in turn performs $j$ fit operations of the regressor - until either a convergence criterion is met, or the number of runs $j$ is exhausted.
In our experiments, we set $r=6, j = 20$. 
Running the complete experiment (attacking all 26 datasets, for seven regressors, and six poisoning rates $\epsilon$) results in $26*6*7 = 1092$ calls to the \emph{iTrim} defense.
On a Intel(R) Xeon(R) CPU E7-4860 v2 @ 2.60GHz with 96 cores, this takes about $120$ minutes when parallelizing into 15 separate processes.
Thus, running a single \emph{iTrim} defense takes, on average, $120 / 1092 * 15 = 1.6$ minutes per $6$-core process.
Obviously, this is highly dependent on the regressor's complexity, the size of the dataset, the number of features, and parallelism capabilities of the program code.
Still, this indicates the feasibility of applying the \emph{iTrim} defense in a real-world scenario, where after weeks, months or even years of data gathering, running the \emph{iTrim} incurs negligible additional time overhead.

% \section{Qualitative Evaluation}
% Additionally, we evaluate our defense on the medical use-case of \emph{Warfarin} prediction presented in Section~\ref{s:case_study_warfarin}.
% By poisoning only two percent of the data, we were able to increase the test loss to $150$ percent using the $Flip$ attack, and increase the rate of unacceptably high/low doses by $21$ percent.
% Using the \emph{iTrim} defense, we can mitigate this damage nearly completely. 
% Out of seven regressors we applied on the thusly poisoned \emph{Warfarin} datasets, all were able to recover the unpoisened MSE - i.e. the MSE when trained on a non-poisoned dataset.
% The increase in the rate of unacceptable doses dropped from $21\%$ back to $0\pm3\%$.
% Table~\ref{tab:warfarin_mae_more} and Section~\ref{s:warfarin_revisited} in the Supplementary Material provide these results in greater detail.

\section{Warfarin Revisited}
\label{s:warfarin_revisited}

\begin{table*}
 \centering
 \caption{Mean absolute error (MAE) of different regression models when poisoned using the \emph{Flip} attack. 
% REMOVED DUE TO SPACE - is already explained in text
%  (C) corresponds to no data poisoning, (P) indicates $2\%$ poison samples and (D) indicates $2\%$ data poisoning and \emph{iTrim} defense. The column \emph{MAE P/C} shows the increase in error due to $2\%$ data poisoning. Similarly, \emph{MAE D/C} shows the relative change in test error after cleaning an attacked dataset using \emph{iTrim}. 
%  The column \emph{Acceptable P/C} shows the decrease of patients receiving an acceptable dose of Warfarin due to the poisoning.
%  The column \emph{Acceptable D/C} shows that this decrease of acceptable doses can be effectively mitigated by \emph{iTrim}.
}
 \input{warfarin_results_more.tex}
\label{tab:warfarin_mae_more}
\end{table*}

In Section \ref{s:case_study_warfarin} we presented the medical use case of predicting the therapeutic Warfarin dose and showed that data poisoning can significantly impact the performance of regressors on the task.
In this section, we will illustrate the empirical results of Section \ref{ss:eval_trim} on the use case of Warfarin dose prediction, where we consider three different scenarios:
First, the (C)lean case. In this scenario, no data poisoning occurs. This case will be used as a baseline for measuring the effects of data poisoning and defence.
Second, the (P)oison case. In this scenario, the attacker introduces $2\%$ poison samples using the \emph{Flip} attack proposed in Section \ref{ss:flip}. No countermeasures are taken.
Third, the (D)efended case. In this scenario the data are poisoned with $2\%$ poison samples like in (P), but \emph{iTrim} is used as a counter measure.
The results for these three scenarios are summarized in Table~\ref{tab:warfarin_mae_more}.

To recapitulate: Warfarin is a blood thinner with a narrow therapeutic window resulting in high medical significance for the correct prediction of the therapeutic Warfarin dose.
The scenarios (C) and (P) have already been presented in Section~\ref{s:case_study_warfarin}.
To summarize: The models used in our evaluation perform comparable to state-of-the-art models and $2\%$ poison samples are sufficient to noticeably increase metrics like the MAE and to decrease the number of patients receiving an acceptable dose of Warfarin by up to $22\%$.

In Table~\ref{tab:warfarin_mae_more} the column 'MAE D/C' provides the factor by which the MAE of a regressor increases, when the dataset is poisoned with $2\%$ poison samples and then defended using \emph{iTrim}.
As we can see, the median is $1.00$, indicating that the damage is mitigated.
The individual values range from $0.98$ to $1.03$, which indicates that where previously \emph{Flip} incurred an increase in MAE of up to $31\%$, the \emph{iTrim} defense reduces this error increase to a tenth.
In summary, the MAE of the tested models in the (D) scenario is approximately the same as in the (C) scenario, meaning the defense successfully eliminates (most) of the negative impact of the poison samples.

The column \emph{Acceptable D/C} gives the percentage by which the number of patients receiving an acceptable Warfarin dose decreases in the (D) scenario compared to the (C) scenario.
The median reduces from $21.07$ in scenario (P) to a median of close to $0$ in scenario (D).
This shows that the number of patients receiving an unacceptable Warfarin dose due to data poisoning is significantly reduced when the \emph{iTrim} defense is employed.
%For some models like MLP and Huber Regression, using \emph{iTrim} can even imp meaning the number of patients receiving an acceptable prediction is even increased.
In summary, we observe that the \emph{iTrim} defense decreases the influence of poison samples.
It results in more patients receiving adequate predictions for their therapeutic Warfarin dose.

\section{Conclusion}
In this paper we introduce a novel data poisoning attack on regression learning as well as a matching defense mechanism.
%We evaluated the current state of literature on data poisoning 
%attacks on regressors. We find that there is only very little research on the subject, and that there is no previous work on poisoning nonlinear regressors.
We show the effectiveness of our proposed attack and defense algorithm in a large empirical evaluation over seven regressors and 26 datasets.
%To mitigate against such data poisoning attacks, we propose \emph{iTrim}.
%, which we evaluate against the 26 datasets, and find that it outperforms existing approaches.
Both attack and defense assume realistic constraints:
The attack is black-box and doesn't assume access to the true dataset, but only a substitute dataset.
%The defense does not require an oracle for the degree (if any) of the poisoning, but is capable of estimating the poison rate.
The defense, on the other hand, does not assume any knowledge of the poisoning rate $\epsilon$, but estimates it using an iterative approach.
%We show that this defense works well both quantitatively over 26 datasets, as well as qualitatively over a running example of predicting Warfarin therapeutic dosage.

\bibliography{paper_biblio}{}
\bibliographystyle{plain}

%\newpage
\section{Supplementary Material}

\begin{figure}[h]
    \centering
    \includegraphics[width=0.82\linewidth]{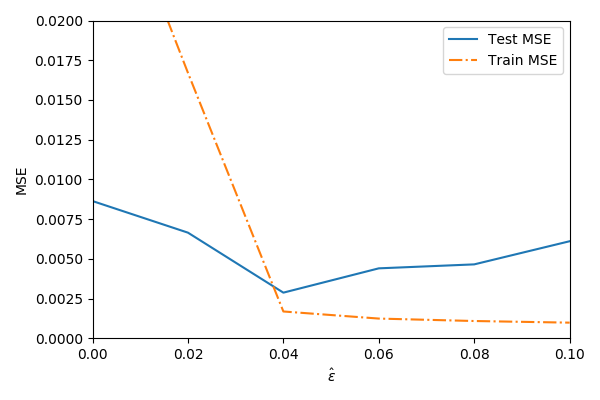}
    \caption{Similarly to Figure~\ref{fig:iTrim_warfarin}, we plot the train vs test error of a KernelRidge regressor on a dataset (%left: \emph{armesHousing}, right:
    \emph{loan}) poisoned with $\hat{\epsilon} = 0.04$.
    We observe that the train loss gives clear indication as to when all poison samples are removed via \emph{iTrim}.}
    \label{fig:iTrim_housing_loan}
\end{figure}
\label{ss:eval_statp}

\subsection{Evaluation of the \emph{StatP} Attack}

\begin{figure}[h]
    \centering
    \includegraphics[width=0.82\linewidth]{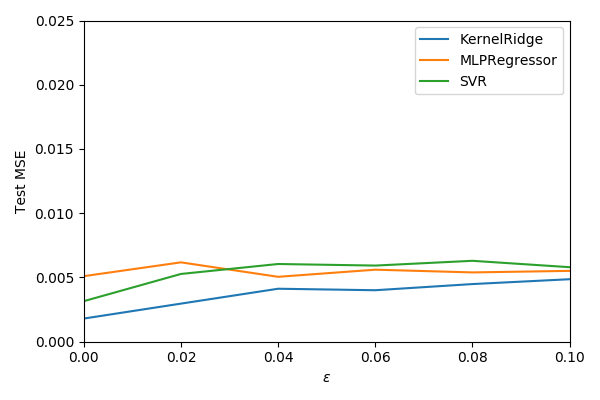}
    \caption{Evaluation of the performance of the \emph{StatP} attack \cite{Jagielski2018} against nonlinear regressors. 
    The \emph{StatP} attack is applied to three nonlinear regressors, averaged over all 26 datasets. 
    The $x$ axis shows the degree of poisoning, while the $y$ axis shows the test MSE. 
    The attack is not effective. 
%     This is because \emph{StatP} is designed for linear regressors and creates out-of-distribution samples. 
    While these do disturb linear regressors, they are easily accommodated by nonlinear models without increasing any test loss. }
    \label{fig:statP}
\end{figure}

\begin{figure*}
    \centering
    \includegraphics[width=0.40\linewidth]{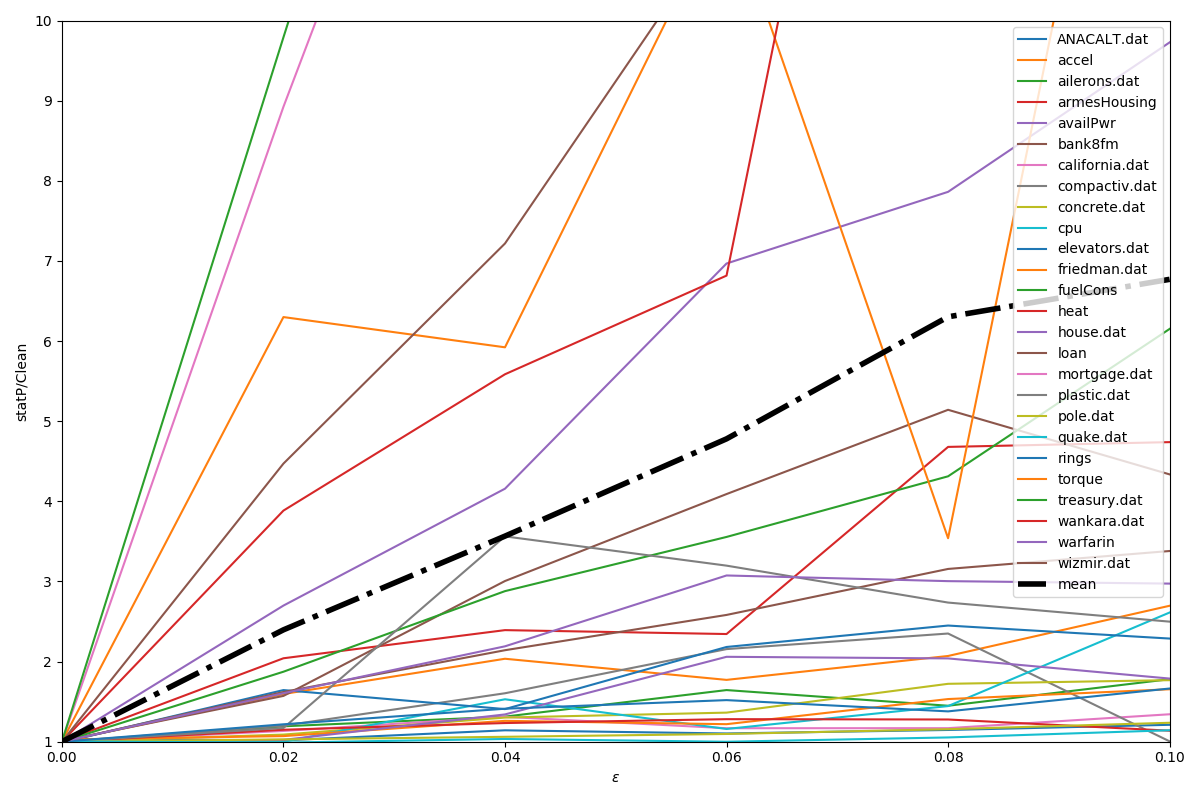}
    \includegraphics[width=0.40\linewidth]{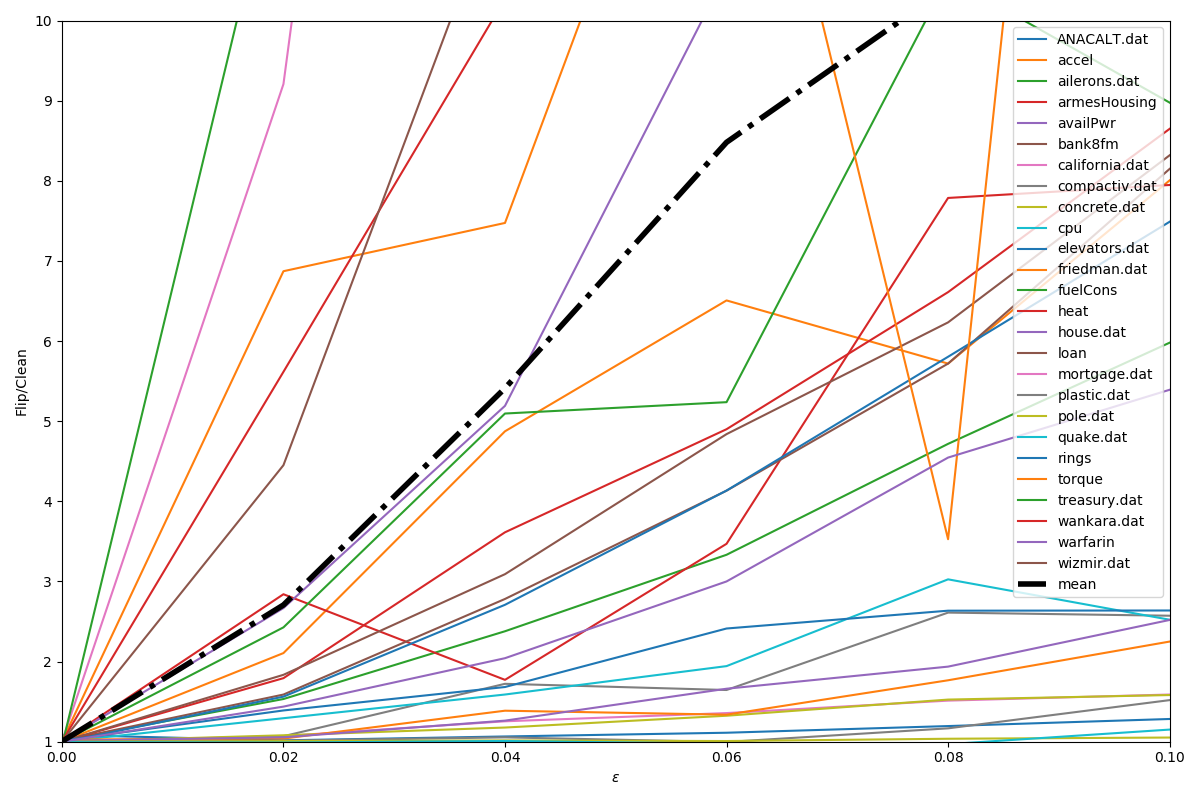}
    \caption{
    \emph{StatP} (left) vs \emph{Flip} (right).
    This plot shows the increase in MAE when using a poisoned dataset instead of a clean dataset during model training.
    The results are averaged over all seven regressors, but displayed individually per dataset.
    For an average over all datasets, refer to Figure~\ref{fig:flip_attack_all}.
    \emph{Left Image:} All 26 datasets when attacked with the \emph{StatP} attack. While the degree of effectiveness varies between datasets, we generally observe a linear correlation between degree of poisoning and increase in test error.
    \emph{Bottom Image:} The same datasets attacked with the \emph{Flip}. Note that this attack consistently outperforms~\emph{StatP}, as seen on the higher MAE.
    }
    \label{fig:attack_results_non_averaged}
\end{figure*}

In this section, we evaluate the performance of existing poisoning attacks on regression learning.
As described in Section~\ref{ss:related_work_attacks}, the only poisoning attack in literature which we can directly apply to regression models in a black box scenario is the~\emph{StatP} attack~\cite{Jagielski2018}.
We implement this attack and evaluate it twice:
First, we apply the attack on the Warfarin dataset and three non-linear models.
% REMOVED DUE TO SPACE
% \footnote{
% % The effectiveness on linear regressors is shown by the authors themselves, and while we can confirm the reported results, we do not re-elaborate them here, but refer to~\cite{Jagielski2018}.
% % }
%  models. % (Figure~\ref{fig:statP}).
Second, we evaluate all 26 datasets against both the \emph{statP} and \emph{Flip} attack (Figure~\ref{fig:attack_results_non_averaged}).
We observe the following:
1.For nonlinear regressors, \emph{StatP} is ineffective. The MAE remains near constant when adding even significant amounts of poison samples (see Figure~\ref{fig:statP}).
    First, when analyzing the poison data created by \emph{StatP}, we find an intuitive explanation for this:
    \emph{StatP} pushes all data points 'to the corners', i.e. to the edge of the feasibility domain.
    While these data do conflict with the 'clean' data for linear regressors, nonlinear models can easily accommodate both poison and clean data, as long as the samples don't overlap in feature-space.
    Second, when evaluatingk both attacks on all 26 datasets, averaging four linear and three non-linear models, we can confirm the above results. \emph{Flip} consistently outperforms \emph{StatP} (c.f. Figure~\ref{fig:attack_results_non_averaged}).

\subsection{Datasets}
The following table lists the datasets we use in our experiments.
The datasets can be obtained at \url{https://sci2s.ugr.es/keel/category.php?cat=reg} and \url{https://github.com/paobranco/Imbalanced-Regression-DataSets} and from~\cite{Jagielski2018}.
\begin{table}[h]
    \centering
    \caption{The 26 datasets we use in our experiments}
    \input{resources/used_datasets.tex}
    \label{tab:datasets}
\end{table}

% \begin{figure}[!htb]
%     \centering
%     \includegraphics[width=0.95\linewidth]{resources/all_reg_trim_vs_itrim_results.png}
%     \caption{Direct comparison of \emph{Trim} and \emph{iTrim}: This plot shows ${\text{MSE}_\text{Trim}}/{\text{MSE}_\text{iTrim}}$. Observe that the quotient is almost always greater than one, indicating a clear dominance of \emph{iTrim} over \emph{iTrim}. Especially for small values of $\epsilon$ \emph{iTrim} shows frequently improvements compared to \emph{Trim}}
%     \label{fig:itim_vs_trim_all}
% \end{figure}

% \begin{figure}
%     \centering
%     \includegraphics[width=\linewidth]{imgs/chap5/trim_vs_itrim.png}
%     \caption{Direct comparison of \emph{Trim} and \emph{iTrim}: This plot shows ${\text{MSE}_\text{Trim}}/{\text{MSE}_\text{iTrim}}$. Observe that the quotient is almost always greater than one, indicating a clear dominance of \emph{iTrim} over \emph{iTrim}.}
%     \label{fig:itim_vs_trim}
% \end{figure}

\end{document}

%% file: warfarin_results.tex
% ===================================================================================================
% \begin{tabular}{lcccc}
% \toprule
% \thead{Model} &  \thead{Clean\\Error} &  \thead{Error \\when \\poisoned} &   \thead{Decrease of \\ acceptable \\ dosages} \\
% \midrule
% Elastic Net     &   8.52 &  11.07 &          21.25\% \\
% Huber Reg. &   8.40 &   8.46 &          -1.09\% \\
% Kernel Ridge    &   8.41 &  10.98 &             21.07\% \\
% Lasso          &   8.49 &  11.20 &            22.31\% \\
% MLP   &  10.05 &  12.33 &             11.41\% \\
% Ridge Reg.         &   8.49 &  10.99 &             22.34\% \\
% SVR            &  10.85 &  12.52 &             14.13\% \\
% \hline
% Median         &   8.49 &  11.07 &             21.07\% \\
% \bottomrule
% \end{tabular}
\begin{tabular}{lcccc}
\toprule
\thead{Model} &  \thead{Clean \\ Error} &  \thead{Pois. \\ Error} &  \thead{Increase \\ MAE} &  \thead{Decrease \\ accep. Dosages} \\
\midrule
Elastic Net     &   8.52 &  11.07 &     1.30 &        21.25\% \\
Huber Reg. &   8.40 &   8.46 &     1.01 &        -1.09\% \\
Kernel Ridge    &   8.41 &  10.98 &     1.31 &        21.07\% \\
Lasso          &   8.49 &  11.20 &     1.32 &        22.31\% \\
MLP   &  10.05 &  12.33 &     1.23 &        11.41\% \\
Ridge Reg.         &   8.49 &  10.99 &     1.29 &        22.34\% \\
SVR            &  10.85 &  12.52 &     1.15 &        14.13\% \\
\hline
Median         &   8.49 &  11.07 &     1.29 &        21.07\% \\
\bottomrule
\end{tabular}
% ===================================================================================================

%% file: warfarin_results_more.tex
% \begin{tabular}{lrrrrrr}
% \toprule
% {} &  MAE C &  MAE P &  MAE P/C &  MAE D/C &  Dec W20 P/C &  Dec W20 D/C \\
% Model          &        &        &          &          &              &              \\
% \midrule
% Lasso          &   8.49 &  11.20 &     1.32 &     0.99 &        22.31 &         1.88 \\
% ElasticNet     &   8.52 &  11.07 &     1.30 &     1.06 &        21.25 &         0.82 \\
% Ridge          &   8.49 &  10.99 &     1.29 &     1.00 &        22.34 &         3.99 \\
% HuberRegressor &   8.40 &   8.47 &     1.01 &     1.01 &        -1.37 &        -3.01 \\
% MLPRegressor   &  10.11 &  12.46 &     1.23 &     0.94 &        16.34 &       -10.46 \\
% SVR            &  10.85 &  12.52 &     1.15 &     1.08 &        14.13 &         6.88 \\
% KernelRidge    &   8.41 &  10.98 &     1.31 &     1.00 &        21.07 &         0.53 \\
% \hline
% Median         &   8.49 &  11.07 &     1.29 &     1.00 &        21.07 &         0.82 \\
% \bottomrule
% \end{tabular}

\begin{tabular}{lrrrrrrr}
\toprule
Model &  MAE C &  MAE P &  MAE D &  MAE P/C &  MAE D/C &  Accbl. P/C &  Accbl. D/C \\
\midrule
Elastic Net     &   8.52 &  11.07 &   8.51 &     1.30 &     1.00 &        21.25 &        -0.82 \\
Huber Reg. &   8.40 &   8.46 &   8.41 &     1.01 &     1.00 &        -1.09 &        -0.82 \\
Kernel Ridge    &   8.41 &  10.98 &   8.41 &     1.31 &     1.00 &        21.07 &         0.53 \\
Lasso          &   8.49 &  11.20 &   8.49 &     1.32 &     1.00 &        22.31 &         0.00 \\
MLP   &  10.05 &  12.33 &   9.86 &     1.23 &     0.98 &        11.41 &        -2.80 \\
Ridge Reg.          &   8.49 &  10.99 &   8.51 &     1.29 &     1.00 &        22.34 &         1.06 \\
SVR            &  10.85 &  12.52 &  11.19 &     1.15 &     1.03 &        14.13 &         2.90 \\
\hline
Median         &   8.49 &  11.07 &   8.51 &     1.29 &     1.00 &        21.07 &         0.00 \\
\bottomrule
\end{tabular}

%% file: resources/used_datasets.tex
\begin{tabular}{llrr}
\toprule
{} &            Name &  features &        n \\
\midrule
0  &     ANACALT.dat &       7.0 &   4052.0 \\
1  &           accel &      22.0 &   1732.0 \\
2  &    ailerons.dat &      40.0 &  13750.0 \\
3  &    armesHousing &     248.0 &   1460.0 \\
4  &        availPwr &      49.0 &   1802.0 \\
5  &         bank8fm &       8.0 &   4499.0 \\
6  &  california.dat &       8.0 &  20640.0 \\
7  &   compactiv.dat &      21.0 &   8192.0 \\
8  &    concrete.dat &       8.0 &   1030.0 \\
9  &             cpu &      12.0 &   8192.0 \\
10 &   elevators.dat &      18.0 &  16599.0 \\
11 &    friedman.dat &       5.0 &   1200.0 \\
12 &        fuelCons &      88.0 &   1764.0 \\
13 &            heat &      30.0 &   7400.0 \\
14 &       house.dat &      16.0 &  22784.0 \\
15 &            loan &     202.0 &   5000.0 \\
16 &    mortgage.dat &      15.0 &   1049.0 \\
17 &     plastic.dat &       2.0 &   1650.0 \\
18 &        pole.dat &      26.0 &  14998.0 \\
19 &       quake.dat &       3.0 &   2178.0 \\
20 &           rings &      10.0 &   4177.0 \\
21 &          torque &      95.0 &   1802.0 \\
22 &    treasury.dat &      15.0 &   1049.0 \\
23 &     wankara.dat &       9.0 &   1609.0 \\
24 &        warfarin &     177.0 &   5528.0 \\
25 &      wizmir.dat &       9.0 &   1461.0 \\
\bottomrule
\end{tabular}